\documentclass{bmvc2k}

\usepackage{amsmath}
\usepackage{amssymb}
\usepackage{bbm}
\usepackage{bm}
\usepackage{upgreek}
\usepackage{algorithm}
\usepackage{algorithmic}
\usepackage{multicol}

\title{Facial Age Estimation by Deep Residual Decision Making }

\addauthor{Shichao Li}{http://vsdl.ust.hk/}{1}
\addauthor{Kwang-Ting Cheng}{http://vsdl.ust.hk/}{1}

\addinstitution{
	Hong Kong University of Science and Technology\\
	Kowloon, Hong Kong
}

\runninghead{Li, Cheng}{Facial Age Estimation by Deep Residual Decision Making}

\def\etal{\emph{et al}\bmvaOneDot}

\begin{document}

\maketitle

\begin{abstract}
Residual representation learning simplifies the optimization problem of learning complex functions and has been widely used by traditional convolutional neural networks. However, it has not been applied to deep neural decision forest (NDF). In this paper we incorporate residual learning into NDF and the resulting model achieves state-of-the-art level accuracy on three public age estimation benchmarks while requiring less memory and computation. We further employ gradient-based technique to visualize the decision-making process of NDF and understand how it is influenced by facial image inputs. The code and pre-trained models will be available at \url{https://github.com/Nicholasli1995/VisualizingNDF}.

\end{abstract}

\section{Introduction}
\label{sec:intro}

Facial age estimation is a task that aims to infer the chronological age from digital facial images, which is challenging due to large appearance variation and varying growing patterns in different life periods~\cite{survey}. Divide-and-conquer is a strategy that solves a complex problem by dividing the problem space and tackle the sub-problems instead. This strategy is inherently used by traditional decision tree based models~\cite{tree_detector, tree_regressor, randomforest} and has been applied on facial age estimation~\cite{randomforest}. Past works using decision trees~\cite{tree_detector, tree_regressor, randomforest} usually adopt hand-crafted features in the hard splitting functions and utilize a heuristic training scheme, limiting their representation learning power. Recent works extend traditional decision trees into deep neural decision forest (NDF)~\cite{NDF, Depth, DRFs} by using soft splitting functions and enables decision tree with deep representation learning ability. 

In another parallel line of research, the architecture of deep convolutional neural networks (CNN) keeps evolving~\cite{VGG,resnet,densenet} to boost model performance on large-scale image classification problem. Among them residual learning~\cite{resnet} was an influential breakthrough by hypothesizing that learning residuals could ease the parameter optimization for very deep neural networks, which is now popularly adopted in CNN architectures~\cite{resdenoise, H_module}. Despite a rigorous mathematical analysis is still beyond our touch, a seminal work by Li~\etal~\cite{loss_ls} visualized the loss function landscape with and without residual learning, which intuitively demonstrated that residual learning did make the optimization problem easier. We believe it's a natural step to use residual learning in other deep models. However, despite its popularity in traditional CNN, residual learning has not yet been attempted for NDF to the best of our knowledge.

Understanding the learned representation and inference process of deep computer vision models is arousing growing interests due to its close connection with the model's trustworthiness~\cite{car}. Gradient-based visualizing~\cite{saliency, invert} and explainable representation learning~\cite{interpret, inter_tree} have already been tried, yet most works focus on image classification task and traditional CNN, leaving regression problems and other deep models less explored.  

In this work we take the step to use residual learning when optimizing the soft decision functions of NDF and apply it on age estimation problem. We also try to better understand the inference process of NDF by computing saliency maps based on the routing probabilities. Finally, considering its remarkable performance~\cite{NDF, DRFs} yet lower popularity compared to traditional CNN, we provide an easy-to-use implementation based on Python and PyTorch to encourage the community to consider NDF for other vision tasks. In summary, our main contributions are twofold: 

\begin{itemize}
	\item[1.] We employ residual learning to learn the complex soft decision functions of NDF for the first time. The trained model achieves state-of-the-art accuracy on facial age estimation task while consumes less memory and computation.
	\item[2.] We are the first to apply network visualization technique on NDF to obtain insightful observations during its inference process.	
\end{itemize}

\begin{figure*}
	\begin{center}
		\includegraphics[width=1\linewidth]{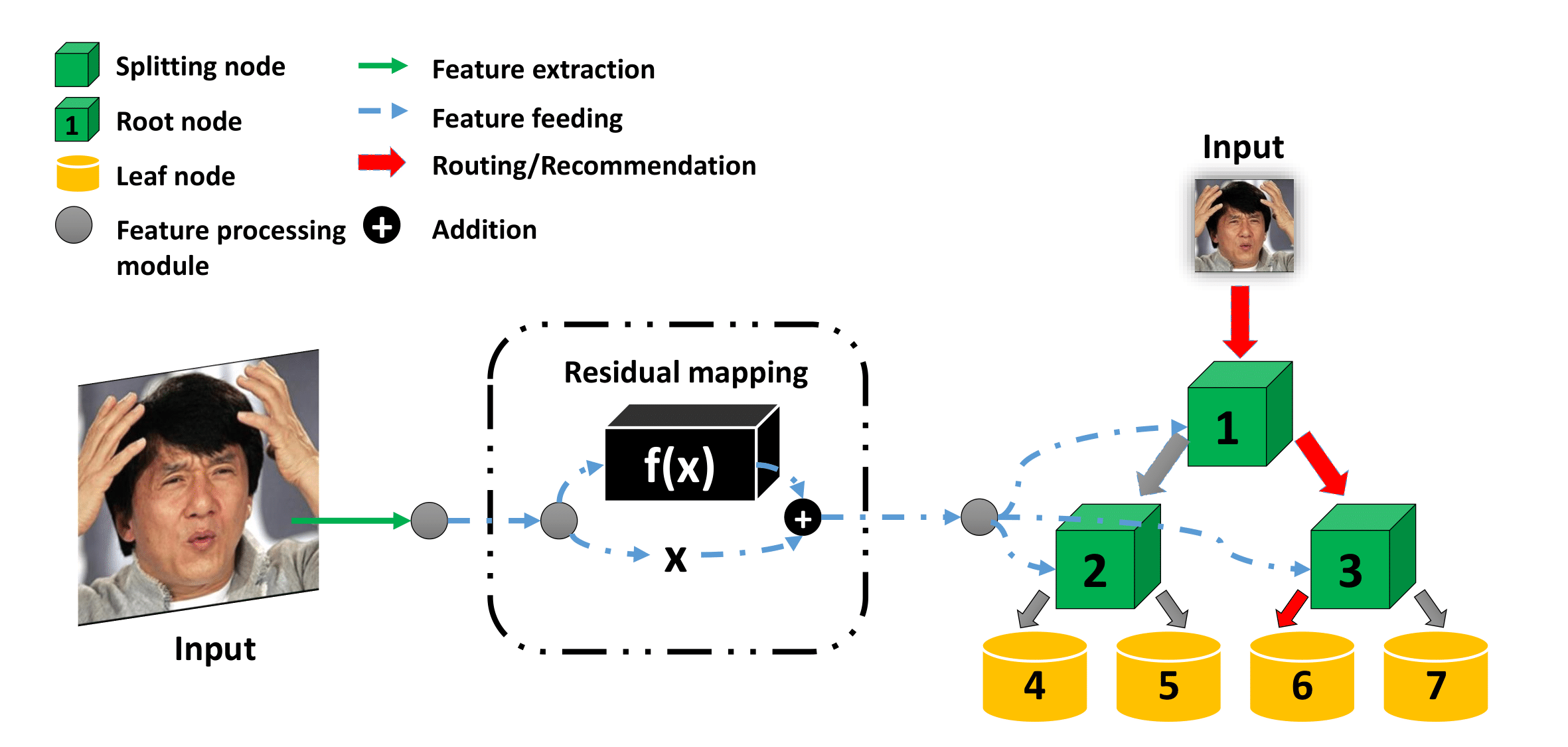}
	\end{center}
	\caption{Illustration of deep residual neural decision forest (RNDF). Input images are routed (red arrows) by the splitting nodes and arrive at the prediction given by leaf nodes. Deep representation is extracted from the input and sent (blue arrows) to each splitting node for decision making. Redisual learning is incorporated in the feature extraction process to help the optimization of complex soft decision functions. Here only one tree of depth two is drawn for simplicity.}
	\label{idea}
\end{figure*}



\section{Related Work}
\textbf{Divide-and-conquer for age estimation.} To reduce the difficulty of facial age regression, past works tried to split the data space and learn multiple regressors~\cite{HumanPerform} or use traditional random forests~\cite{randomforest}. Recent works combine end-to-end deep representation learning with soft decision tree~\cite{NDF,Depth,DRFs} and achieved state-of-the-art accuracy~\cite{DRFs}. However, conventional feature extractor~\cite{VGG} was employed in~\cite{DRFs} which resulted in inefficient parameter usage and residual learning was not attempted. 

\textbf{Residual learning.} After being proposed in~\cite{resnet}, residual learning has been widely adopted in CNN to solve various problems~\cite{resdenoise,VQA,H_module}. Nevertheless, it has not yet been used for NDF, which is a another type of deep model different from traditional CNN.

\textbf{Network visualizing and explainable AI.}. Network visualization techniques~\cite{saliency, invert} have been used to understand the learned representation in CNN, yet not applied for NDF where a different type of inference is conducted via decision making. These works also focused on image classification and not on a regression problem like age estimation. Recent efforts tried to build explainable CNN model~\cite{interpret} and organize it by decision tree~\cite{inter_tree}. Our work is different from them since we do not add extra loss function in the training phase. Our previous extended abstract \cite{li2019visualizing} was targeted for image classification, and here we extend it to facial age regression. 

\section{Methodology}
\subsection{Residual Neural Decision Forest}
A deep neural decision forest (NDF) is an ensemble of deep neural decision trees. Each tree consists of splitting nodes and leaf nodes. In general each tree can have unconstrained topology but here we specify each tree as full binary tree for simplicity. We index the nodes sequentially with integer $i$ as shown in Figure~\ref{idea}. A splitting node $\mathcal{S}_{i}$ is associated with a recommendation (splitting) function $\mathcal{R}_{i}$ that extracts deep features from the input $\mathbf{x}$ and gives the recommendation score (routing probability) $s_{i} = \mathcal{R}_{i}(\mathbf{x})$ that the input is recommended (routed) to its left sub-tree. We denote the unique path from the root node to a leaf node $\mathcal{L}_{i}$ a computation path $\mathcal{P}_{i}$. Each leaf node stores one real-valued prediction vector $\mathbf{p}_{i}$ that represents the "answer" given by it. To get the final prediction $ \mathbf{P}$, each leaf node contributes its prediction vector weighted by the probability of taking its computation path as  
\begin{equation}
\mathbf{P} = \sum_{i\in\mathcal{N}_{l}}w_{i}\mathbf{p}_{i}
\end{equation}
and $\mathcal{N}_{l}$ is the set of all leaf nodes.
The weight can be obtained by multiplying all the recommendation scores given by the splitting nodes along the path. Assume the path $\mathcal{P}_{i}$ consists of a sequence of $q$ splitting nodes and one leaf node as $\{\mathcal{S}_{i_1}^{j_1}, \mathcal{S}_{i_2}^{j_2}, \ldots, \mathcal{S}_{i_q}^{j_q}, {\mathcal{L}_{i}}\}$, where the superscript for a splitting node denotes to which child node to route the input. Here $j_m = 0$ means the input is routed to the left child and $j_m = 1$ otherwise. Then the weight can be expressed as
\begin{equation}
w_{i} = \prod_{m=1}^{q}(s_{i_m})^{\mathbbm{1}(j_m = 0)}(1 - s_{i_m})^{\mathbbm{1}(j_m = 1)}
\end{equation} 

Note that the weights of all leaf nodes sum to 1 and the final prediction is hence a convex combination of all the prediction vectors of the leaf nodes. In addition, we assume the recommendation functions mentioned above are differentiable and parametrized by $\boldsymbol{\uptheta}_{i}$ at node $i$. Then the final prediction is a differentiable function with respect to all the parameters which we omit above to ensure clarity. A loss function defined upon the final prediction can hence be minimized with gradient descent.

Similar to previous works~\cite{NDF, DRFs}, we use deep CNN to extract features from the input and assign each splitting node to one neuron of the last fully-connected layer, where sigmoid function is used to compute the final recommendation scores. Specifically,
\begin{equation}
s_{i} = \mathcal{R}_{i}(\mathbf{x}) =
\sigma(f_i(\mathcal{M}_{n}(...\mathcal{M}_{1}(\mathbf{x}))))
\end{equation}
where $\mathcal{M}_{k}$ is the $k$th feature mapping function represented by one or multiple layers in deep neural networks, $f_i$ is the linear mapping function associated with the assigned neuron in the last fully-connected layer and $\sigma()$ is sigmoid function to enforce a valid range for routing probability. Inspired by the success of residual learning in CNN~\cite{resnet}, we hypothesize that learning residual can help the learning of recommendation functions, where we specify the feature mapping functions as
\begin{equation}
\mathcal{M}_{k}(\mathbf{x}) = \mathbf{x} + \mathcal{H}_{k}(\mathbf{x})
\end{equation}
and we call the NDF incorporated with residual learning as residual neural decision forest (RNDF).
 
For a regression problem where the dataset contains $N$ labeled instances $\mathbb{D}=\{\mathbf{x}_i,\mathbf{y}_i\}_{i=1}^{N}$, we directly use the squared loss function, 

\begin{equation}
L(\mathbb{D}) =\frac{1}{2} \sum_{i=1}^{N}\lvert\lvert\mathbf{P}_{i}-\mathbf{y}_{i}\rvert\rvert^2
\end{equation}

The network parameters are optimized by gradient descent keeping the leaf node prediction vectors fixed. For a splitting node $\mathcal{S}_i$, we denote nodes in its left and right sub-trees as node sets ${\mathcal{S}_i^l}$ and ${\mathcal{S}_i^r}$, respectively. We denote the probability of recommending the input $\mathbf{x}$ to a leaf node $\mathcal{L}_{i}$ as $\mathbb{P}(\mathcal{L}_{i}|\mathbf{x})$. The gradient of loss function with respect to the recommendation score 
 $\frac{\partial L(\mathbb{D})}{\partial{s_i}}$ is computed as,
\begin{equation}
\frac{\partial L(\mathbb{D})}{\partial{s_i}} = \sum_{t = 1}^{N}(\mathbf{P}_{t} - \mathbf{y}_{t})^T
(\frac{\mathbf{A}_l}{s_{i}}
- \frac{\mathbf{A}_r}{(1-s_{i})})
\label{gradient}
\end{equation}
where $\mathbf{A}_l = \sum_{\mathcal{L}_j\in \mathcal{S}_i^l}\mathbb{P}(\mathcal{L}_{j}|\mathbf{x}_t)\mathbf{p}_{j} $ and $\mathbf{A}_r = \sum_{\mathcal{L}_j\in \mathcal{S}_i^r}\mathbb{P}(\mathcal{L}_{j}|\mathbf{x}_t)\mathbf{p}_{j} $. This gradient is back-propagated to the former layers to optimize layer parameter with gradient descent.

The leaf node prediction vectors are optimized by leaf node update rule keeping the network paramters fixed. There are different leaf node update rules available~\cite{Depth, DRFs} and we adopt the one~\cite{DRFs} with theoretically guaranteed loss reduction performance. Here a covariance matrix $\Sigma_i$ is used at each leaf node to specify prediction uncertainty and a Gaussian distribution is assumed with the prediction vector used as the mean. The prediction vector and covariance matrix can be updated jointly as
\begin{equation}
\mathbf{p}_j^{T+1} = 
\frac{\sum_{t=1}^{N}\mathbf{y}_t\zeta_j(\mathbf{p}_j^{T},\bm{\Sigma}_j^{T}|\mathbf{y}_t)
}{\sum_{t=1}^{N}\zeta_j(\mathbf{p}_j^{T},\bm{\Sigma}_j^{T}|\mathbf{y}_t)}
\label{update_p}
\end{equation}
\begin{equation}
\bm{\Sigma}_j^{T+1} = 
\frac{\sum_{t=1}^{N}
	(\mathbf{y}_t - \mathbf{p}_j^{T+1})(\mathbf{y}_t - \mathbf{p}_j^{T+1})^{\textrm{T}}
	\zeta_j(\mathbf{p}_j^{T},\bm{\Sigma}_j^{T}|\mathbf{y}_t)
}{\sum_{t=1}^{N}\zeta_j(\mathbf{p}_j^{T},\bm{\Sigma}_j^{T}|\mathbf{y}_t)}
\label{update_sigma}
\end{equation}
where the weighting factor $\zeta_j(\mathbf{p}_j^{T},\bm{\Sigma}_j^{T}|\mathbf{y}_t)$ is computed as 
\begin{equation}
\zeta_j(\mathbf{p}_j^{T},\bm{\Sigma}_j^{T}|\mathbf{y}_t) =
\frac{\mathbb{P}(\mathcal{L}_{j}|\mathbf{x})
	p_j(\mathbf{y}_t)}
{\sum_{i=1}^{\mathcal{N}_{l}}\mathbb{P}(\mathcal{L}_{i}|\mathbf{x})
	p_i(\mathbf{y}_t)}
\end{equation}
and $p_j(\mathbf{y}_t)$ is the probability density given by the assumed leaf node Gaussian distribution
\begin{equation}
p_j(\mathbf{y}_t) = \frac{1}{\sqrt{(2\pi)^k|\bm{\Sigma}_j^{T}|}}exp[-\frac{1}{2}
(\mathbf{y}_t - \mathbf{p}_j^{T})^{\mathrm{T}}
({\bm{\Sigma}_j^{T}})^{-1}
(\mathbf{y}_t - \mathbf{p}_j^{T})]
\end{equation} 
The gradient descent and leaf node update rule are carried out alternately and the training process is depicted by Algorithm~\ref{training}.

\begin{algorithm}
	\caption{Training algorithm for RNDF}
	\label{alg1}
	\begin{multicols}{2}
		\begin{algorithmic}[1]
			\REQUIRE
			training set $\mathbb{D}=\{\mathbf{x}_i,\mathbf{y}_i\}_{i=1}^{N}$, trainable network parameters $\bm{\theta}$ and leaf node parameters $\{\mathbf{p}_j, \bm{\Sigma}_j\}$, SGD batch number $B_n$
			\STATE Initialize $\bm{\theta}$  and $\{\mathbf{p}_j, \bm{\Sigma}_j\}$  randomly. 
			\WHILE {Not converge}
			\STATE $B_i = 0$, fix $\{\mathbf{p}_j, \bm{\Sigma}_j\}$
			\WHILE {$B_i < B_n$}
			\STATE Select a random batch from $\mathbb{D}$ 
			\STATE Update $\bm{\theta}$ by SGD (Eqn.~\ref{gradient})
			\STATE $B_i = B_i + 1$
			\ENDWHILE
			\STATE Select a random batch and update $\{\mathbf{p}_j, \bm{\Sigma}_j\}$ (Eqn.~\ref{update_p} and Eqn.~\ref{update_sigma})
			\ENDWHILE
		\end{algorithmic}
	\end{multicols}
	\label{training}
\end{algorithm}

\subsection{Decision Saliency Map}
To understand how the input can influence the decision-making of this model, we take the gradient of the routing probability with respect to the input and name it {\itshape decision saliency map (DSM)},

\begin{equation}
DSM = \frac{\partial s_{i}}{\partial \mathbf{x}}
\end{equation}

The definition of DSM is inspired from the past gradient-based network visualization technique~\cite{saliency}, but is unique since the past work focus on traditional CNN and image classification while here the saliency map is computed for NDF and age regression. NDF conducts inference by decision-making and saliency maps are more meaningful in this scenario. In experiment we trace one computation path for each input and compute DSMs for each splitting node on the path. Multiple paths are available for the same input since there are multiple leaf nodes and trees. We take the path that contributes most, i.e., the path whose weight $w_i$ is the largest.

\begin{table}
	\begin{center}
		\begin{tabular}{|l|c|c|c|c|}
			\hline
			Method & Year& Morph & FG-NET & CACD\\
			\hline\hline
			AGES~\cite{Method:Ages} &2007&8.83/46.8\%&6.77/64.1\%&-/-\\
			LARR~\cite{LARR} &2008&-/- & 5.07/68.9\% & -/- \\
			IIS-LDL~\cite{LDL} &2010&-/-&5.77/-&-/-\\
			Rank~\cite{Method:rank} &2010&6.49/49.1\%&5.79/66.5\%&-/-\\
			MTWGP~\cite{MTWGP}  &2010&6.28/52.1\%&4.83/72.3\%&-/-\\
			CAM~\cite{CAM} &2011&-/- & 4.12/73.5\% & -/- \\
			OHRank~\cite{OHRank}  &2011&6.07/56.3\%&4.48/74.4\%&-/-\\
			CPNN~\cite{LDL} &2013&-/-&4.76/-&-/-\\
			CA-SVR~\cite{CASVR}  &2013&5.88/57.9\%&4.67/74.5\%&-/-\\
			DIF~\cite{HumanPerform} &2015&-/-&4.80/74.3\%&-/-\\
			Human Workers~\cite{HumanPerform} &2015&6.30/51\%&4.70/69.5\%&-/-\\
			DLA~\cite{DLA} &2015&4.77/63.4\%&4.26/-&-/-\\
			Rothe \etal~\cite{Rothe_2016_CVPR} &2016&3.45/-&5.01/-&-/-\\
			DEX~\cite{DEX} &2016&3.25/-&4.63/-&4.785/-\\
			dLDLF~\cite{dLDLF} &2017&3.02/81.3\% & -/- & 4.734/- \\
			ARN~\cite{ARN} &2017&3.00/- & -/- & -/- \\
			DRFs~\cite{DRFs} &2018&\textbf{2.91}/82.9\% & \textbf{3.85}/\textbf{80.6}\% & 4.637/-  \\
			RNDF (Ours) &2019&2.97/\textbf{83.2}\% & 3.87/76.1\% & \textbf{4.595}/- \\
			\hline
		\end{tabular}
	\end{center}
	\caption{Mean absolute error (MAE) and cumulative score (CS) of different methods are reported as MAE/CS. Some previous works do not report both metrics.}
	\label{comparison}
\end{table}

\section{Experiments}
\subsection{Datasets}
We employ three public datasets for experimental evaluation:
\begin{itemize}
	\item[1.] FG-NET~\cite{FGNET} contains 1002 images taken from 82 subjects at different ages. These images have large variation in terms of pose, illumination and facial expression. We conduct leave-one-out cross validation as previous works~\cite{DEX, DRFs}. In each experiment images from 81 subjects are used as the training set and the model is validated on the images from the remaining one subject. The experiments are repeated 82 times to validate on every subject and the final results are averaged.
	\item[2.] MORPH~\cite{MORPH} contains 55134 annotated images from more than 13000 individuals of different races. We follow the previous work~\cite{DRFs} by selecting 5475 images\footnote{The list of selected images were released by the previous work at \url{https://github.com/shenwei1231/caffe-DeepRegressionForests/blob/master/morph_setting1.list}} and randomly choose 80\% of them for training and validate on the remaining images. The experiments are repeated 5 times and the averaged results are reported. 
	\item[3.] CACD~\cite{CACD} is a large challenging dataset containing 166417 images collected from 2000 celebrities from Internet. The celebrities are grouped into three subsets: the training set containing 145275 images from 1,800 celebrities, the testing set that has 10517 images from 120 celebrities and the validation set having the remaining images from 80 celebrities. We train our model on the training subset and report its performance on the test subset. 
\end{itemize}

\begin{table}
	\begin{center}
		\begin{tabular}{|l|c|c|c|}
			\hline
			Method & MAE~(CACD) & Model size & FLOPs\\
			\hline\hline
			DRFs~\cite{DRFs} &4.637 & 539.4MB & 16G \\
			RNDF (Ours) &\textbf{4.595} & \textbf{112.4MB} & \textbf{4G} \\
			\hline
		\end{tabular}
	\end{center}
	\caption{Comparison of model size and FLOPs.}
	\label{comparison2}
\end{table}
\subsection{Implementation details}
\textbf{Preprocessing}. We use facial landmarks to locate the face region and eliminate in-plane face rotation. Only Morph dataset does not officially provide facial landmarks and we use OpenCV and Dlib for face detection and alignment. When the detector fails we manually crop and the face region. Finally we resize all images to 256 by 256 pixel and normalize the images based on computed mean and standard deviation of three color channels before feeding them to the model. The training data is augmented by random horizontal flipping with probability 0.5 and random cropping so that the final spatial size of the inputs is 224 by 224. For testing images we only conduct central cropping.

\begin{figure}
	\centering
	
	\begin{center}
		\includegraphics[width=1\linewidth]{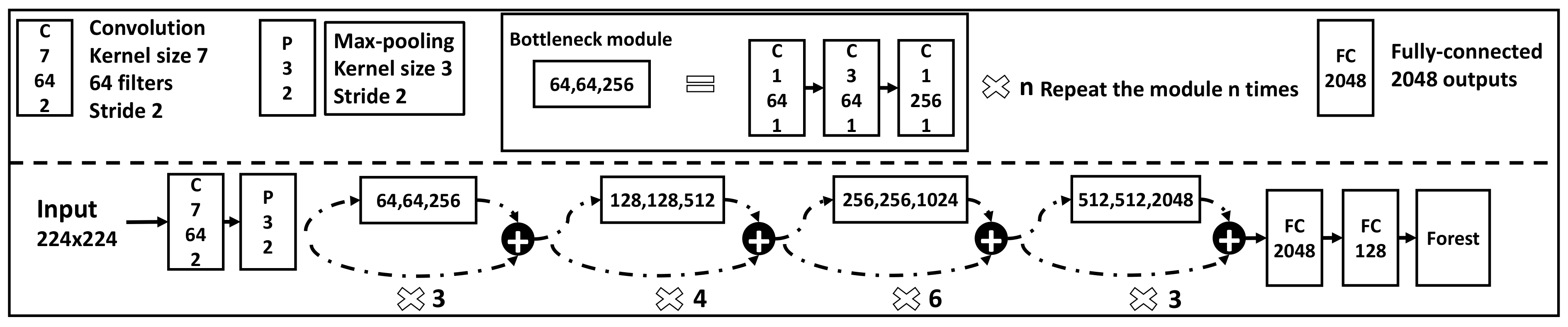}
	\end{center}
	
	\caption{Detailed architecture of RNDF. The input goes through four types of residual bottleneck blocks and two fully-connected layers, whose output (after sigmoid activation) are sent to the forest for decision making.} 
	\label{modelarc}
\end{figure}

\textbf{Model Architecture.} The detailed model architecture is shown in Fig.~\ref{modelarc}, where a Resnet50-like~\cite{resnet} architecture is adopted and two fully-connected layers are used whose final outputs are activated by sigmoid function and sent to the forest to give the final prediction. 

\textbf{Hyper-parameter setting.} For fair comparison with the previous work~\cite{DRFs}, we only use a forest of 5 trees and each one is of depth 6.  A batch size of 50 is used for back-propagation to train the network parameters and we did not notice significant performance discrepancy with a batch size range from 15 to 100. We update the leaf node prediction vectors after every 50 batches of network parameter update ($B_n = 50$), where 500 samples are randomly drawn for each time. Each leaf node update will run 20 iterations of the update rule.

\textbf{Training settings.} SGD optimizer is used with momentum set as 0.9 and the initial learning rate is 0.5. We train 40 epochs in every experiment for FG-NET, 100 epochs for Morph and 8 epochs for CACD. The learning rate is halved when model training gets stuck in plateau using the scheduler provided by PyTorch. Detailed settings can be found in our released code.
\subsection{Results}
We use two widely used metrics to evaluate the accuracy of our model. Mean absolute error (MAE) is the average absolute error over the testing set and cumulative score (CS) is the portion of testing images whose test error is smaller than a threshold. Following previous works\cite{OHRank, DRFs} we use a threshold of 5 years. 

The model accuracy on the three benchmarks is shown in Table~\ref{comparison} and compared with previous works. Our model achieves state-of-the-art level accuracy on all of the benchmarks and noticeably good performance on the largest dataset CACD. The memory and computation efficiency of the model is compared with the previous most accurate model\cite{DRFs} in Table~\ref{comparison2}, where our model has a memory saving of 4.8 times and computation saving of 4 times.

\begin{figure}
	\centering

	\begin{center}
		\includegraphics[width=1\linewidth]{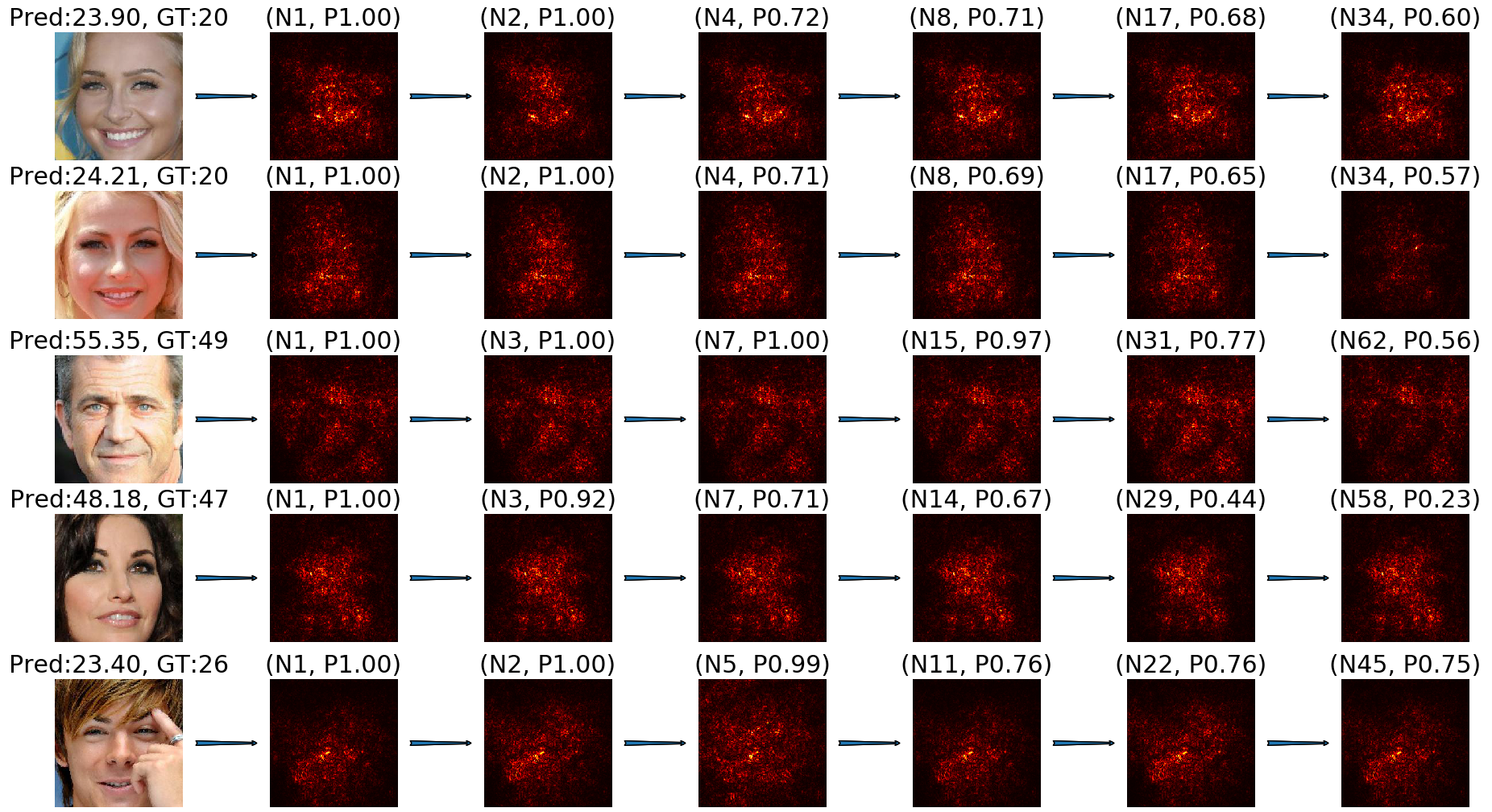}
	\end{center}

	\caption{Decision saliency maps for CACD. Each row gives the decision-making process of one image, where the left-most image is the input and the others are DSMs along the computation path of the input. Each DSM is computed by taking derivative of the routing probability $s_i$ with respect to the input image. Model prediction and ground truth are given above the input image as (Pred, GT). (Na, Pb) means the input arrives at splitting node a with probability b during the decision-making process.}
	\label{results_CACD}
\end{figure}

\begin{figure}
	\begin{center}
	\begin{center}
		\includegraphics[width=1\linewidth]{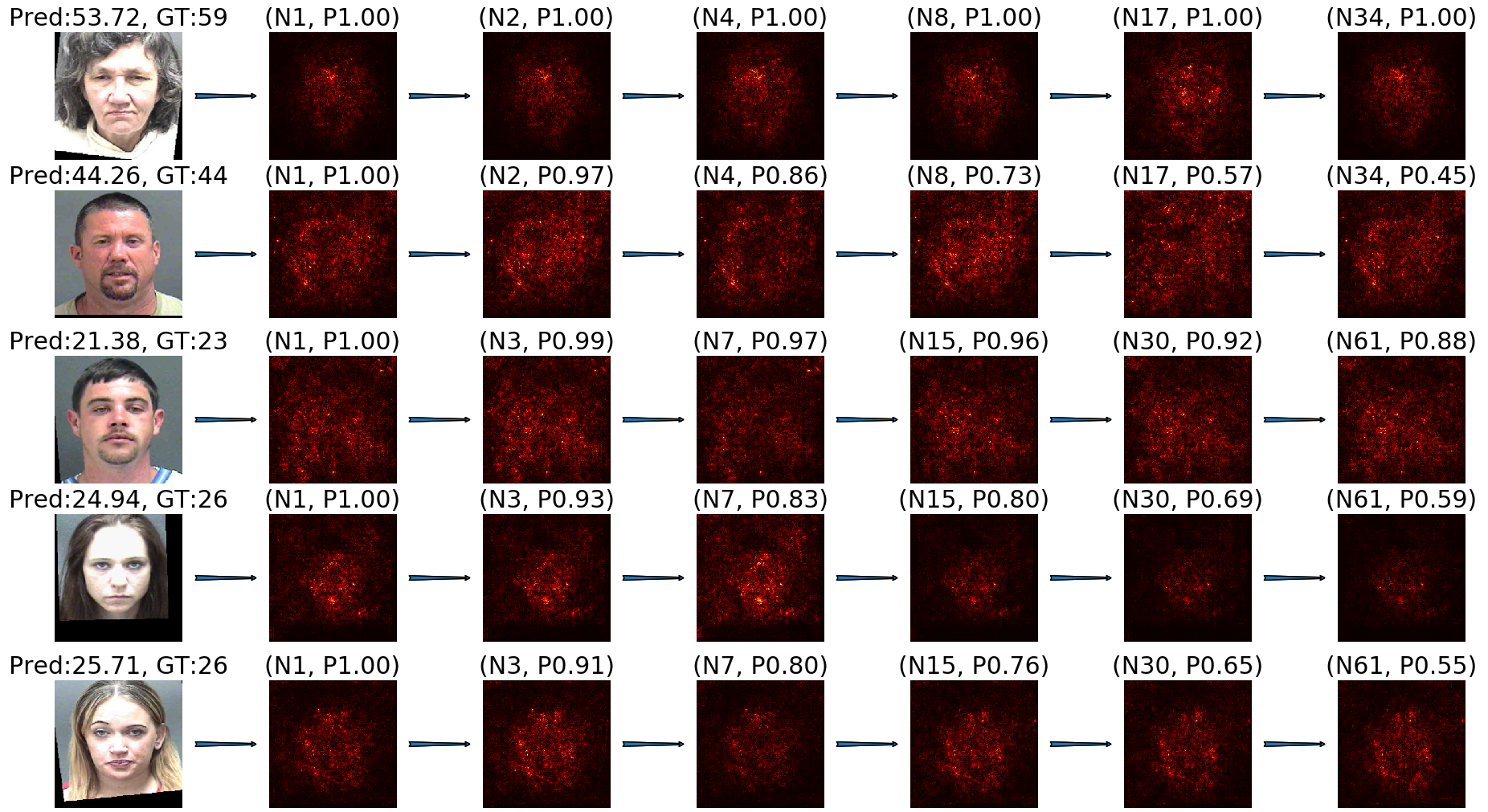}
	\end{center}
	\end{center}
	\caption{Decision saliency maps for Morph using the same annotation style as Figure~\ref{results_CACD}.}
	\label{results_morph}
\end{figure}

The DSMs for Morph and CACD datasets are shown in Fig.~\ref{results_CACD} and Fig.~\ref{results_morph}, respectively. It can be seen that for different input, the model learns to make decision based on different regions. Note that the decision making is usually based on the skin region and the model learns to ignore irrelevant texture (the hair region and background) for most cases. Another observation is that the splitting nodes along the path usually "look at" similar facial regions, and the reason maybe all the splitting nodes are associated with the same fully-connected layer in the model. Finally, we intentionally use different face sizes in the input (face region occupies a larger portion for CACD input than Morph) to show that the model is not sensitive to pre-processing. 
\section{Conclusion}
In this work we employ residual learning, a successful technique validated by traditional convolutional neural networks, in deep neural decision forest (NDF) to learn the soft decision functions. Our model achieves state-of-the-art accuracy on large facial age estimation dataset, requires less memory and is more computationally efficient. We also apply network visualization technique on NDF to obtain deeper understanding of the decision-making process of this model.  

The are several angles for future research. Firstly, all the trees are limited to full binary trees and all the splitting nodes are restricted to one fully-connected layer in this study. It's an interesting question whether one can learn different tree topology or associate splitting functions to different network layers. In that case the model can be more flexible and can combine different information from different layers. Secondly, we only compute saliency maps for this model and one can also employ other network visualization techniques for NDF. Finally, the network architecture in this work is a first trial and we expect future works to further improve the accuracy or compress the model size.

\bibliography{reference}
\end{document}